\documentclass{article}

\usepackage[final]{neurips_2021}

\usepackage[utf8]{inputenc} 
\usepackage[T1]{fontenc}    
\usepackage{hyperref}       
\usepackage{url}            
\usepackage{booktabs}       
\usepackage{amsfonts}       
\usepackage{nicefrac}       
\usepackage{microtype}      
\usepackage{xcolor}         
\usepackage[pdftex]{graphicx}

\title{Controlled Cue Generation for Play Scripts}

\author{\stepcounter{footnote}\vspace{1mm}Alara Dirik$^{}$\thanks{Equal contribution. Author ordering determined by a coin flip.}\quad Hilal Donmez$^{}$\footnotemark[2] \quad Pinar Yanardag\\
Boğaziçi University\\
Istanbul, Turkey\\
{\tt\small \{alara.dirik, hilal.donmez\}@boun.edu.tr} \\ {\tt\small yanardag.pinar@gmail.com}
}

\begin{document} 
\maketitle

\begin{abstract}
In this paper, we use a large-scale play scripts dataset to propose the novel task of theatrical cue generation from dialogues. Using over one million lines of dialogue and cues, we approach the problem of cue generation as a controlled text generation task, and show how cues can be used to enhance the impact of dialogue using a language model conditioned on a dialogue/cue discriminator. In addition, we explore the use of topic keywords and emotions for controlled text generation. Extensive quantitative and qualitative experiments show that language models can be successfully used to generate plausible and attribute-controlled texts in highly specialised domains such as play scripts. Supporting materials can be found at: \url{https://catlab-team.github.io/cuegen}.
\end{abstract}

\section{Introduction}

Script generation for theater plays involves the automatic generation of a sequence of lines of dialogue and cues that are coherent as a whole. While story and plot generation are relatively popular tasks, play and movie script generation remains a largely unexplored problem. In this paper, we focus on the generation of theatrical cues from character dialogue lines. A theatrical cue can be described as an informative text that is not spoken dialogue. It can be a trigger for an action, an informative description of the stage, thoughts of the characters or body language intended to amplify the effect of the play. Cues are highly variable in context and can range from sound effects, lighting changes, the movement of characters on stage, moods, thoughts, and reactions via silent gestures. The following example illustrates how a cue is used to direct a character's action on stage and add to their spoken lines.

\begin{quote}
    {\textbf {JOHN: I don't know what to do anymore.}} \\
     {\textbf {(JOHN turns around and leaves.)}}
\end{quote}
In addition to describing the actions of the characters, cues also describe the interaction between them, such as the following example:
\begin{quote}
    {\textbf {LIZZIE: How do you…? (Putting things together:) No . . .}} \\
     {\textbf {POYDRAS: But you also have her eyes.}}\\
          {\textbf {LIZZIE: (Weeps. Realizes she is looking at her father. Takes a moment.)
}}
\end{quote}
Theatrical cues play an important role in screenplays by bridging the gap between the audience and the actors. They can set the tone of the scene and conversations and bring otherwise mundane conversations to life. Therefore, cue generation is a useful and valuable tool for playwrights to explore different creative avenues and inspire actors and actresses to present their performances with great impact. Another common use case for theatrical cues is to modernise and reinterpret old plays without changing the dialogue. In our work, we thoroughly investigate this use case by generating plausible cues based on the original dialogue lines. To this end, we have collected over 1500 play scripts with various topics, containing a total of 775,000 lines of dialogue and over 277,000 cues. To the best of our knowledge, we are the first to propose the novel task of generating cues from dialogues in plays. 

In this work, we introduce a new task and use large-scale transformer-based language models trained on large text corpus for controlled text generation.  Controlling the attributes of the generated text, such as specific topics or sentiments, remains difficult without fine-tuning the models for each attribute separately. To address this issue, we explore cue generation using the preceding dialogue and propose a cue/dialogue discriminator using the PPLM framework proposed by \cite{Dathathri2020PlugAP}. We also explore other extensions such as emotion-based and topic-based text generation.

\section{Related Work}

\subsection{Text generation}

Text generation is a very popular NLP task where deep neural networks are widely used, with sequence-to-sequence (seq2seq) (see \cite{Sutskever2014SequenceTS}) with attention (see \cite{Luong2015EffectiveAT}) among the most popular models. Generative adversarial networks (GAN) \ and autoencoders (see \cite{Wang2018SentiGANGS,Hu2017TowardCG}) have also been used to generate text conditioned on specific attributes. These works focus on training generative models and variational autoencoders for style transfer, which rely on learning disentangled latent representations for style and content.

Most of the work on text generation in recent years has been based on the transformer architecture (see \cite{Vaswani2017AttentionIA, elikyilmaz2020EvaluationOT, Hu2017ControllableTG, Keskar2019CTRLAC}), which has enabled training large-scale language models (LMs) on very large datasets and significantly improved the state-of-the-art in natural language processing, as \cite{Radford2018ImprovingLU} shows. BERT by \cite{devlin2018bert} and GPT-2 by \cite{Radford2019LanguageMA} are among the most successful transformer-based language models. Recent studies have used BERT for conditional text generation, employing a large pre-trained language model to generate text conditioned on intent labels (see \cite{xia2020cg}). Similarly, \cite{Sheng2020TowardsCB, Prabhumoye2020ExploringCT, Ziegler2019FineTuningLM} have conducted studies on using GPT-2 to generate text with controlled attributes and biases. However, these approaches are often not useful in practice as they require the model to be fine-tuned for each specific attribute separately. In our work, we focus on plug-and-play approaches and generate text by steering pre-trained language models towards acquiring the target attributes.

\subsection{Story Generation}

Previous research in story generation such as \cite{clark2018neural} mostly focuses on using recurrent neural networks (RNNs) and long short term memory units (LSTMs) for text generation. However, RNNs have difficulties in generating longer and coherent texts (see \cite{bahdanau2014neural, sutskever2014sequence, cho2014learning}), hence other works such as \cite{martin2018event} aim to provide different semantic representations for story generation.

\cite{martin2018event} proposed dividing the automated story generation task into two subtasks: successive generation of events (event2event) and generation of human-readable sentences from events (event2sentence). The event2event model generates successive events by extracting semantic information from each sentence and the event2sentence model translates the generated events into human-readable sentences. Controllable story generation (see \cite{peng2018towards}) is another text generation method that uses an analyzer consisting of supervised classifiers and rule-based keyword extractors to extract control factors from story corpus and a generator that generates stories with an RNN conditioned on the control factors. While this approach can be used to generate stories that reflect the user's intent, a separate model needs to be trained for each new intent or control factor.

Interactive story generation is another research area where various machine learning methods have been proposed (see \cite{riedl2013interactive}). Interactive story generation enables users to influence or direct stories with their inputs. \cite{brahman2020cue} focused on the task of interactive story generation, where the user provides mid-level sentence abstractions in the form of cue phrases to the model during the generation process. \cite{akoury2020storium} proposed another story generation system called STORIUM, where human authors query a model for suggested story continuations and edit them.

\subsection{Dialogue Systems}
The rise of deep learning based Natural Language Understanding (NLU) and Natural Language Generation (NLG) methods has significantly improved the performance of dialogue systems. Dialogue systems typically consist of two modules: an NLU module to extract information from user queries and an NLG module to produce relevant responses and start new dialogues. Since dialogue generation directly depends on the performance of the NLU approach used, it is critical to understand the user intent correctly. \cite{vanzo2019hierarchical}) tried to solve this problem by proposing a hierarchical multitask NLU architecture that creates a domain-independent and rich semantic representation of the user input. This approach aims to encode the structure of the user input along with the actions and arguments it contains via a self-attention mechanism, seq2seq BiLSTM encoders, and CRF tagging layers. Once the user intent is extracted, a conditional text generation method such as a conditional variational autoencoder (see \cite{d2020conditioned}) can be used to generate user-intent dependent responses.

\subsection{Play Script Generation}
The vast majority of previous work on creative text generation focuses on song lyrics generation, story generation (see \cite{Luo2019LearningTC, Jain2017StoryGF}), and movie plot and script generation (see \cite{Zhu2020ScriptWriterNS, Martin2018EventRF, Mangal2019LSTMVG}), while theater play script generation is explored to a much lesser extent. \cite{htgaa} trained a character-level RNN model on theater play scripts to generate entire plays and stage directions. However, previous work on creative text generation mainly investigates how to generate coherent, reasonable, and diverse stories and scripts.  Since creating labeled datasets with the desired attributes is time-consuming and labor-intensive, this work limits the controllability of the generated texts to coarse-grained sentiments (e.g. positive, negative intent). Hence, fine-grained controllable play script generation remains an unexplored topic to the best of our knowledge. 
 
More recently, \cite{rosa2020theaitre} proposed THEaiTRE, a mixed framework that consists of generative language models and hierarchical generation approaches that use text summarization and machine translation methods. THEaiTRE finetunes a pre-trained GPT-2 model \cite{Radford2019LanguageMA} on a small dataset of formatted theater and movie scripts in English and Czech. Moreover, this work proposes to generate a new training dataset by cross-translating between Czech and English to overcome the limited amount of training data. However, it is not possible to evaluate the performance of this approach as the dataset, experimental results and generated play scripts have not been released.

\section{Dataset}
We have collected 1511 English-language play scripts with over 775,000 lines of dialogue and over 277,000 cues on a variety of themes including \textit{Comedy}, \textit{Romance}, \textit{Satire}, and \textit{Greek}. The collected play scripts are scraped from the Playscripts website\footnote{https://www.playscripts.com} and usually include the title of the play, production notes, background information on the characters, and the play itself. 

A play script is a highly structured text consisting of one or more acts defined by elements such as rising action, climax, and resolution. Each act consists of six or more scenes, with each scene containing conversations between 2-4 characters. While acts represent a broader storyline of interrelated events, a scene usually represents actions that take place in one place and time, and are delineated from the next scene by a curtain, a blackout, or a brief emptying of the stage. Therefore, conversations within a scene are often separate from the preceding scenes and take place between different characters.

\begin{figure}
\centering
\begin{minipage}{.4\textwidth}
  \centering
  \includegraphics[width=1\linewidth]{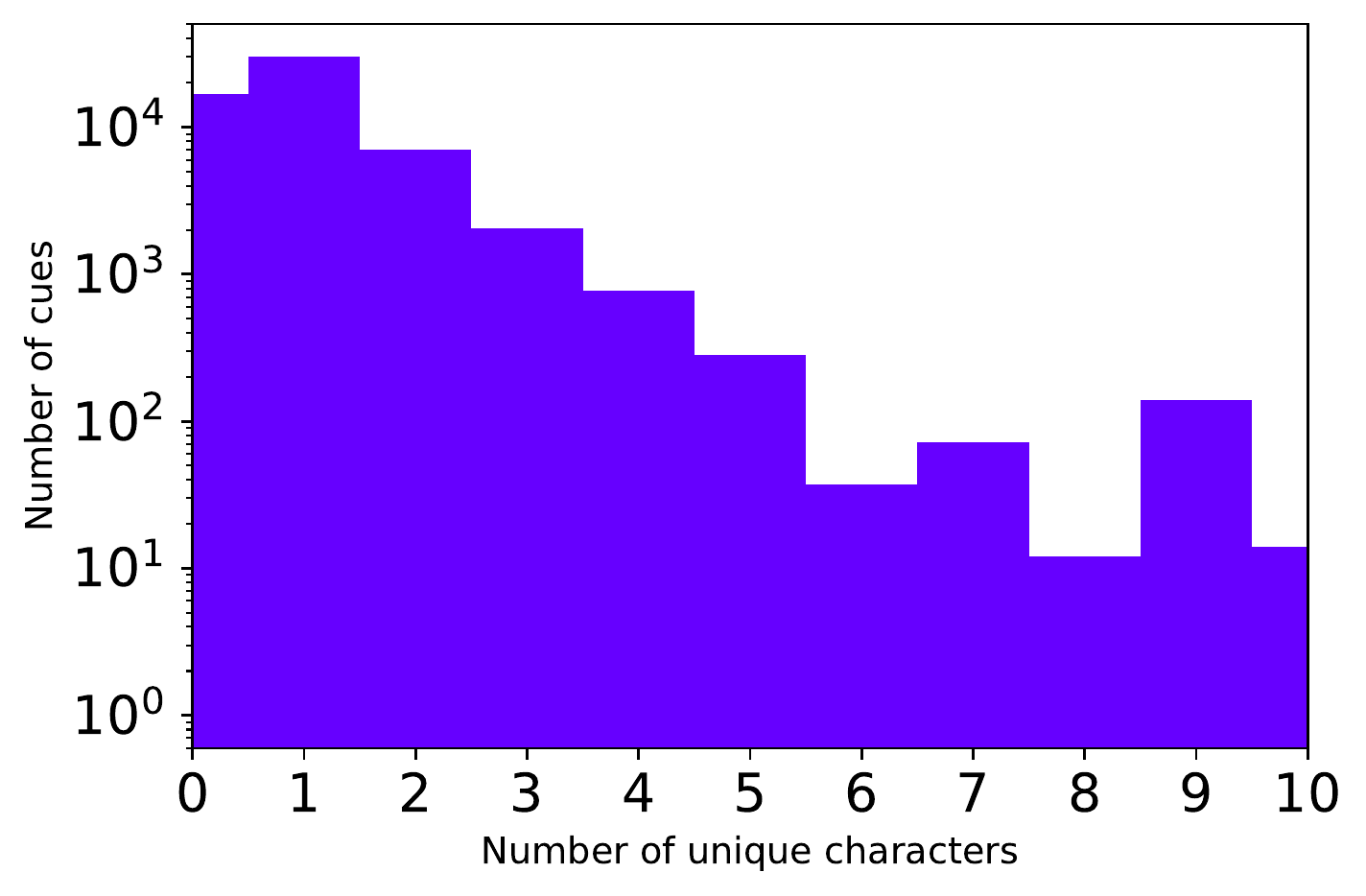}
  \label{fig:chars}
\end{minipage}%
\begin{minipage}{.4\textwidth}
  \centering
  \includegraphics[width=1\linewidth]{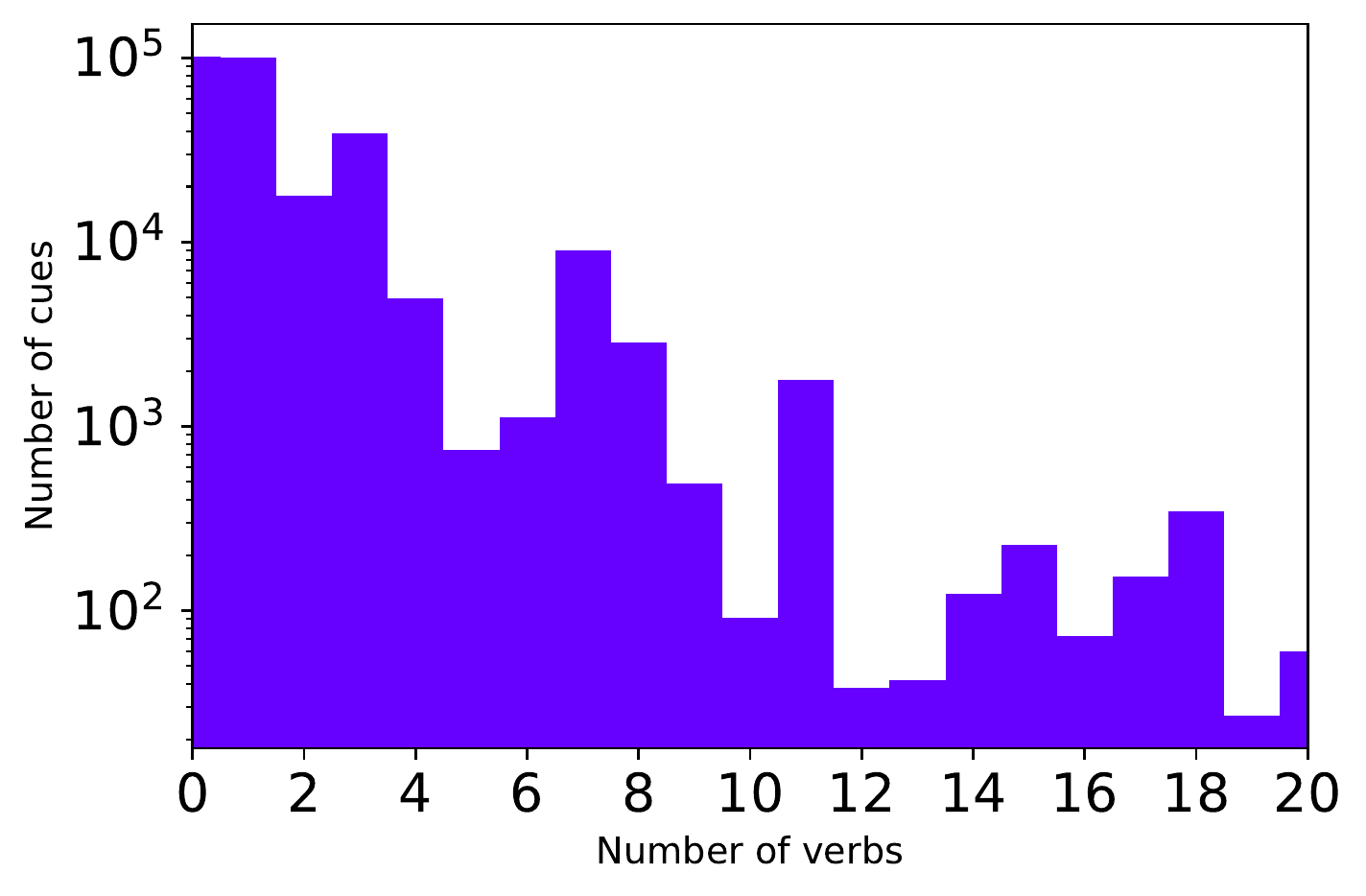}
  \label{fig:verbs}
\end{minipage}
\caption{Histogram of number of unique character names (left). Histogram of number of verbs in cues (right).}
\label{fig:chars_verbs}
\vskip -0.2in
\end{figure}

Each scene in a play script consists of lines of dialogue and cues. In all play scripts, dialogue lines start with capitalized character names and cue lines are placed in parentheses. In our work, we pre-process raw scripts by eliminating pages that do not contain at least one line of dialogue and one line of cues. Cues are not meant to be spoken aloud by characters, and their lengths are highly variable. They contain stage directions, stage descriptions, and character descriptions that are essential to understanding the scenes, as well as the mood, feelings, and thoughts of the characters conveyed through silent expressions. Cues are valuable tools for actors to communicate with the audience and convey spoken lines/dialogue in a myriad of different ways. In addition, cues are often used to modernise and/or reinterpret plays without changing the dialogue. For example, a cold greeting as opposed to a friendly greeting can say a lot about the relationship between two characters. In addition to indicating the feelings of the characters, cues can also be stage directions such as:
   
 \begin{quote}
    {\textbf {(Silence as ROLAND exits stage left.)}} \\
     {\textbf {(LOWELL looks toward the stage right door.)}}\\
    {\textbf {(GRAHAM runs into the bathroom, stage right. He begins to vomit loudly. The knocking becomes even more persistent.)}}\\
\end{quote}

A manual review of the dataset revealed that stage directions and scene changes make up a small portion of the dataset. To distinguish stage directions and scene changes from the rest of the cues, we counted the number of cues containing the word \textit{stage} and found that 11K out of 227K cues contain the keyword stage. The number of character names in the cues varies widely. As shown on the left in Figure \ref{fig:chars_verbs}, some cues contain no character names, while some cues contain up to 10 characters. Cues can also describe actions that characters are supposed to perform (e.g. "Suddenly jumps up from the chair"). To analyze the categories of these actions, we examined the number of verbs that appear in the cues. The right side of Figure \ref{fig:chars_verbs} shows that some cues contain no action, while some of them can have up to 20 actions.

\section{Methodology}
\label{sec:methodology}

Plug and Play Language Models (PPLM) aim to leverage large pre-trained language models (LM) to generate attribute controlled text without fine-tuning or re-training the models. In the context of our work, controllable generation refers to modeling the conditional likelihood of generated text $p(x|a)$, where $a$ denotes desired controllable attribute(s) such as emotion, topic, sentence type/intent and $x$ is the generated sample. PPLM plugs an attribute model $p(a|x)$ together with a base generative model $p(x)$ (GPT-2) and sample from the resulting conditional likelihood $p(x|a) \propto p(a|x)p(x)$. Therefore, it effectively creates a conditional generative model on the fly from any given attribute model, where the attribute models are either in the form of a bag-of-words (BoW) or a discriminator with a single learned layer, without any further training of the underlying base LM. 

\begin{figure}
\centering
\includegraphics[width=.7\textwidth]{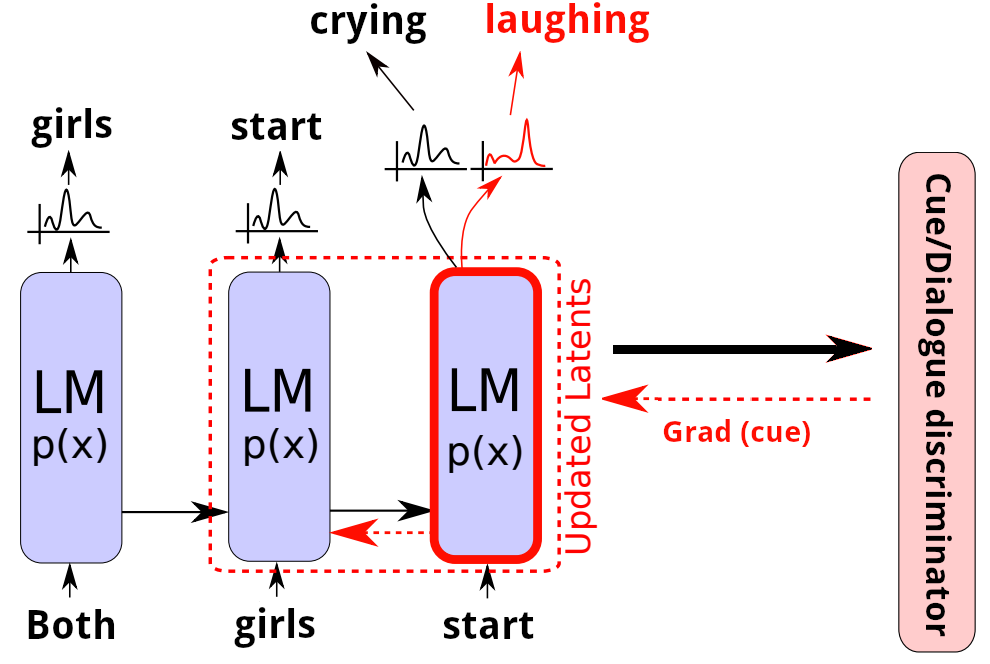} 
\caption{An illustration of the PPLM approach with cue/dialogue discriminator. Figure is modified from  \cite{Dathathri2020PlugAP}.}
\label{fig:pplm}
\vskip -0.2in
\end{figure} 

The PPLM method uses GPT-2 medium as the base LM, which is a left-to-right autoregressive model that generates one token at a time, using the preceding text as input. Given a sequence of tokens or preceding text $\left\{x_{0}, \cdots, x_{n-1}\right\}$, transformer based LMs compute the unconditional probability of the resulting sequence $p(X)$ for all succeeding token candidates:
\begin{equation}
p(X)=\prod_{i=1}^{n} p\left(x_{i} \mid x_{0}, \cdots, x_{i-1}\right)
\end{equation}
Moreover, the GPT-2 architecture uses the hidden representation $H_t$ to generate $x_{t+1}$, given $x_t$. In order to steer the output of the LM, PPLM shifts the hidden representations $H_t$ towards the sum of two gradients at each generation step $t$: towards the higher log-likelihood of attribute $a$ under the conditional attribute model $p(a|x)$, and towards the higher log-likelihood of the base LM $p(x)$. Thus, the shifted hidden representation $\left(H_{t}+\Delta H_{t}\right)$ leads to a distribution of generated text that is more likely to contain the selected attribute(s). As in the original PPLM experiments, we initialize $\Delta H_{t}$ to zero and update it with gradients from the attribute model that measures the closeness between the generated text and the desired attribute such as a topic, emotion, intent.
 
Furthermore, $\Delta H_{t}$ is updated to minimize the KL divergence between the output distribution of the modified and unmodified language models to ensure fluency. In addition to minimizing KL divergence, post-norm fusion is performed similarly to \cite{Stahlberg2018SimpleFR} to bind the generated text to the unconditional $p(x)$ LM distribution. 
 
We note that the baseline PPLM framework uses only seven manually generated lists of topic words and a sentiment discriminator trained on the IMDB movie reviews dataset (see \cite{Maas2011LearningWV}), which is insufficient for our task. Therefore, we use PPLM as our base framework and train a cue/dialogue sentence type discriminator to condition the generation towards cues (see Figure \ref{fig:pplm}). In addition to the cue/dialogue sentence type discriminator, we introduce and experiment with two other attribute models: an automated topic modeling module and an external multi-label emotion classifier, Deep-Moji (see \cite{Felbo2017UsingMO}), for controlled text generation. While the dialogue/cue discriminator and the topic-based approach aim to generate appropriate cues, the emotion classifier is used to steer the generated text towards the emotion label of the input text.  We describe the details of the three attribute models we use in Sections \ref{sec:cue}, \ref{sec:lda} and \ref{sec:emo}.

\subsection{Controlled Generation using Cue/Dialogue Discriminator}
\label{sec:cue}
We train a binary cue/dialogue discriminator using 10\% of our dataset, where the input sentences $x$ are tagged with their corresponding labels $y$. The discriminator consists of a single-layer classifier that predicts the target label. Based on the sentence type given as input by the user and the classifier prediction, PPLM shifts the activations towards the higher log-likelihood of either the dialogue lines or cues as specified by the user.

\subsection{Controlled Generation using LDA} 
\label{sec:lda}
Unlike PPLM, where the lists of topic keywords are created manually, we create word lists by automatically extracting topics using Latent Dirichlet Allocation (LDA) (see \cite{Blei2003LatentDA}). To this end, we create a cue corpus and model it as a distribution of 10 topics. We use the trained LDA model to extract topic keywords and automatically generate topic keyword lists. A target topic selected by the user is then used to steer the language generation process to maximize the log-likelihood of the extracted target topic keywords and generate cues with the target topic.  

\subsection{ Controlled Generation with Emotions} 
\label{sec:emo}
Since plays contain a wide range of emotions, and not just positive or negative sentiments, we train an emotion classifier using DeepMoji, a sentiment model that predicts the emoji label of input sentences (see \cite{Felbo2017UsingMO}). We use DeepMoji to predict the emojis corresponding to the given lines, and then map the predicted emojis to a subset of emotions from Plutchik's Wheel of Emotions (see \cite{Plutchik1980EmotionAP}). We then use the input sentences (dialogue lines) and their corresponding emotion labels to train an emotion classifier. The trained classifier is used to steer the generation towards the target emotion label and does not necessarily generate cues.

 \begin{table} 
 \caption{Model output samples for GPT-2+FT and PPLM+CueDisc methods.}
\label{table_examples}
\centering
\smallskip\begin{tabular}{|p{3.5cm}|p{3.5cm}|p{3.5cm}|}
\hline
Input & GPT-2+FT & PPLM+CueDisc \\\hline
 OWEN. Yeah. He’s awesome. & MEG. Yeah. He is. He has the biggest smile on earth… OWEN. Oh. & WADSWORTH. You think he likes me, too? OWEN. (Laughs, but then gets serious: )\\
 \hline
 VAL. He’s been looking at me. & STEWART. Who is he? VAL. I don't know, never seen him.   & (VAL stares back at JOE, who looks confused.) VAL. I said, “What’s his problem?” \\
 \hline
JACKIE. Okay, I can do this. Thanks. & RENEE. You can do this? You can do this! & (JACKIE starts to walk away. The other girls, all laughing, stop her.) \\
\hline
CAL. My mother is dead. & MADELINE. She died of a drug overdose. CAL. That’s a lie. & (She pulls back the sheet.) I don’t want to know how. \\
\hline
STEWART. Yes! Yes, it is. This is it! & The moment. You know what I
mean. We have to get out of here right now! & MISS BLAINE. It’s not.
STEWART. (Suspiciously:) What do you mean? \\
\hline

\end{tabular}
\end{table}

\subsection{Experimental Setup}
We compare the PPLM-based extensions with fine-tuned GPT-2 and Infilling by Language Modeling (ILM) \cite{Donahue2020EnablingLM} baselines. For the PPLM experiments, we use the GPT-2 345M model fine-tuned on 80\% of our dataset as our base generative model $p(x)$. While the GPT-2 model is not fine-tuned in the original work, the structure and rigid syntax of play scripts require fine-tuning the model to generate plausible dialogues and cues. We use 10\% of the dataset to perform steered inference and test the PPLM approaches, and the remaining 10\% to train the conditional attribute models $p(a|x)$: a binary cue/dialogue classifier, a multi-label emotion classifier, and for topic modeling via LDA to be used in the PPLM experiments. We also perform a simple preprocessing step to insert a white space between the punctuation and the alphanumeric characters. For the ILM experiments, we first divide our dataset into training, validation and test sets in a ratio of 80-10-10 and create infilling examples. To do this, we divide the dataset into successive triples of lines, where the lines can be either dialogues or cues in any order. We also randomly mask paragraphs, sentences, n-grams, and words with a masking probability of 3\% each, resulting in a marginal token masking rate of 15\%. For fine-tuning, we insert a bos token $<BOS>$ at the beginning of each scene and an eos token $<EOS>$ at the end of each scene to mark the beginning and end of different conversations. We filter the training and test datasets to only include consecutive dialogue-cue-dialogue triplets and use the start and end dialogue lines as input during inference.

\begin{itemize}
    \item \textbf{GPT-2+ FT }: Given a line of dialogue as input, we use a GPT-2 model fine-tuned on our dataset to generate text. \item \textbf{ILM}: ILM enables LMs to infill variable-length spans using both preceding and subsequent text. We follow the same approach proposed in the ILM paper and fine-tune the GPT-2 small model on successive line triples following the order dialogue-cue-dialogue. The second line of the triplet is masked during the training and sampling processes since our goal is to generate cues. Once trained, infilling is performed by using the preceding and succeeding dialogue lines as inputs to the model. 
    \item \textbf{PPLM+LDA}: We extract keywords using LDA and control the generation process based on the topic of the dialogue. 
    \item \textbf{PPLM+CueDisc}: We train a cue/dialogue sentence type discriminator and control the generation process using this classifier. 
    \item \textbf{PPLM+Emotion}: We train a multi-label emotion classifier and steer the generation process to generate text that reflects the target emotion specified by the user.
\end{itemize}

\paragraph {Parameter setup} For the PPLM experiments, we use the official PyTorch implementation published by the authors with modifications and extensions. We use the same parameters as PPLM to fine-tune the GPT-2 model. For all PPLM experiments, we set the step size $\alpha$ to $0.04$, the scaling coefficient for the normalization term $\gamma$ to $1.0$. Additionally, we keep the default values for the KL coefficient $\lambda_{ KL }$ and the gamma scale $\gamma_{gm}$, which are $0.01$ and $0.95$ respectively. The number of update steps $m$ is 1 for all experiments, as we found that a larger number of update steps leads to more deterministic results. For ILM experiments, we train an ILM model with the default fine-tuning parameters specified in the Transformers library (see \cite{Wolf2019HuggingFacesTS}), except that we use a batch size of 24 and a sequence length of 256. For all model experiments, we use a seed value of $0$ and perform inference on a single GPU.

\subsection{Quantitative Results}
We use n-gram similarity and distance metrics (see \cite{Kondrak2005NGramSA}) to measure the similarity of the generated text to our reference cue corpus, which consists of 50,000 cue samples from our training set. We generate 600 samples with each model and determine the top 10 reference cues for each sample that yield the smallest Levenshtein distance to the generated text. The Levenshtein distance is defined as the minimum number of elementary edit operations required to transform one string to another. We then compute the unigram and bigram similarity (LCSR and BI-SIM) for each generated sample and closest reference cue pairs, and report the average similarity over all generated samples. \textbf{PPLM+Emotion} and \textbf{PPLM+LDA} samples are generated using randomly selected target emotions and topics respectively. As shown in Table \ref{table1}, PPLM with the Dialogue/Cue discriminator (denoted as \textbf{PPLM+CueDisc}) achieves the highest LCSR and BI-SIM scores, indicating that \textbf{PPLM+CueDisc}) can successfully generate cues. The \textbf{PPLM+Emotion} and \textbf{PPLM+LDA} approaches achieve the second and third best LCSR and BI-SIM scores respectively. Since the \textbf{PPLM+Emotion} approach aims to generate text solely based on target emotion rather than sentence type (dialogue/cue), the results suggest that the dataset relies heavily on cues to convey target emotions, and that \textbf{PPLM+Emotion} therefore generates cues rather than dialogue lines. The fine-tuned GPT-2 medium model (denoted \textbf{GPT-2+ FT }) and the Infilling by Language Modeling (denoted \textbf{ILM}) have low LCRS and BI-SIM scores, suggesting that they are unable to generate relevant and complex cue structures.

\begin{table} [!h]
\caption{LCSR and BI-SIM scores of the models. PPLM+CueDisc shows the best performance in terms of LCSR and BI-SIM metrics.}
\label{table1}
\centering
\smallskip\begin{tabular}{lll}
\hline
Method & LCSR $\uparrow$ & BI-SIM $\uparrow$\\
\hline
GPT-2+FT & 0.42 & 0.29 \\
ILM & 0.47 & 0.24 \\
PPLM+CueDisc & \textbf{0.72} & \textbf{0.60} \\
PPLM+LDA & 0.68 & 0.55 \\
PPLM+Emotion & 0.69 & 0.57 \\
\hline
\end{tabular}
\end{table}

In addition, we measure the diversity of the text generated by each model by the number of distinct n-grams (normalized by the length of text) as in \cite{Li2016ADO}. We report the Dist-1, Dist-2, and Dist-3 scores for the distinct 1-2-3-grams in Table \ref{distinct}. As can be seen in Table \ref{distinct}, \textbf{PPLM+CueDisc} and \textbf{PPLM+Emotion} are comparable or better than \textbf{GPT-2+ FT } in generating diverse text while \textbf{ILM} and \textbf{PPLM+LDA} perform worst. On closer inspection, we find that some of the extracted cue keywords do not refer to characters, but to stage directions such as scene changes, lighting and sound instructions. Therefore, using the keywords extracted with LDA sometimes leads to the generation of repetitive, non-character related text. Similarly, we note that \textbf{ILM} also tends to generate repetitive text that resembles stage directions. Some examples of scripts generated with (\textbf{GPT-2+ FT } and \textbf{PPLM+CueDisc}) can be found in Table \ref{table_examples}. As can be seen from the examples, the GPT-2+ FT method is capable of generating plausible text, but not necessarily cues. In contrast, our method is able to generate cues with the characters that appear in the input text.
 
\begin{table} 
\caption{Dist-1, Dist-2, Dist-3 scores of the models. }
\label{distinct}
\centering
\smallskip\begin{tabular}{llll}
\hline
Method & Dist-1 $\uparrow$ & Dist-2 $\uparrow$ & Dist-3 $\uparrow$\\
\hline
GPT-2+FT & 0.32 & 0.71 & 0.82 \\
ILM & 0.18 & 0.62 & 0.72 \\
PPLM+CueDisc & 0.25 & 0.69 & 0.80 \\
PPLM+LDA & 0.20 &  0.58 & 0.72 \\
PPLM+Emotion & \textbf{0.34} & \textbf{0.74} & \textbf{0.87} \\
\hline
\end{tabular}
\end{table}

\subsection{Qualitative Results}

We asked 20 human annotators to evaluate the performance of the models based on the coherence of the generated text and the accuracy of the cue generation. To create an evaluation dataset, we selected the best performing PPLM-based approach \textbf{PPLM+CueDisc} with the top competitor approach \textbf{GPT-2+ FT } and generated 50 random examples with each model. We then asked the evaluators to rate the generated examples based on coherence and cue accuracy in a binary manner. In the context of our work, we define coherence as both the independent plausibility of the generated text and the contextual coherence of the generated text with respect to the input sentence. Furthermore, we define cue accuracy as whether or not the text generated by the model contains a cue or not.

\begin{table} [!h]
\caption{Qualitative analysis with 20 human evaluators. The evaluators are asked whether the generated texts contain any cue (cue accuracy) and are coherent.}
\label{qualresults}
\centering
\smallskip\begin{tabular}{lll}
\hline
Method & Cue Acc $\uparrow$ & Coherence $\uparrow$\\
\hline
GPT-2+FT & 32.4 & 66.3 \\
PPLM+CueDisc & \textbf{92.5} & \textbf{69.0} \\
\hline
\end{tabular}
\end{table}

As can be seen from Table \ref{qualresults}, while \textbf{GPT-2+ FT } generates diverse text, it fails to generate cues given an input sentence. Since the majority of the dataset consists of dialogues, it is expected that the \textbf{GPT-2+ FT } approach is biased towards generates dialogues. On the other hand, our method achieves a high cue accuracy score while preserving the overall coherence of the conversation. However, we strongly believe that the coherence of the generated texts can be improved by better preprocessing steps and persona-based discriminators. We leave these ideas for future work.

\section{Conclusion} 
In this paper, we use a large-scale play script dataset and propose the novel task of generating theatrical cues from dialogues. We approach the cue generation problem as a controlled text generation task and use a plug-and-play language model with a cue/dialogue discriminator, LDA-based topic keyword lists, and a multi-label emotion classifier to steer the language model to the desired attributes without re-training the model. Our experiments show that language models can be successfully used to generate plausible and attribute-controlled text in highly specialized domains such as plays. In the future, we plan to explore character and person-based cue and dialogue generation tasks with plug-and-play models.

\section*{Acknowledgements}
This publication has been produced benefiting from the 2232 International Fellowship for Outstanding Researchers Program of TUBITAK (Project No:118c321). We also acknowledge the support of NVIDIA Corporation through the donation of the TITAN X GPU.
\bibliography{biblio}
\bibliographystyle{plainnat}


\end{document}